\documentclass[lettersize,journal]{IEEEtran}
\usepackage{amsmath,amsfonts}
\usepackage{array}
\usepackage[caption=false,font=normalsize,labelfont=sf,textfont=sf]{subfig}
\usepackage{textcomp}
\usepackage{stfloats}
\usepackage{url}
\usepackage{verbatim}
\usepackage{graphicx}
\usepackage{cite}
\usepackage{times}
\usepackage{epsfig}
\usepackage{graphicx}
\usepackage{amsmath}
\usepackage{amssymb}
\usepackage{tikz}
\usepackage{comment}
\usepackage{amsmath,amssymb}
\usepackage{color}
\usepackage[accsupp]{axessibility}
\usepackage{booktabs}
\usepackage{times}
\usepackage{epsfig}
\usepackage{bm}
\usepackage{algorithmicx,algorithm}
\usepackage{multirow}
\usepackage{eqparbox}
\usepackage{color}
\usepackage{float}
\usepackage{makecell}
\usepackage[noend]{algpseudocode}
\usepackage{graphbox}
\usepackage{array,booktabs}

\usepackage{comment}
\usepackage{stmaryrd}
\usepackage{breqn}
\usepackage{caption}
\DeclareMathOperator*{\argmax}{argmax}

\algnewcommand{\LeftComment}[1]{\Statex \(\triangleright\) #1}
\usepackage[capitalize]{cleveref}
\crefname{section}{Sec.}{Secs.}
\Crefname{section}{Section}{Sections}
\Crefname{table}{Table}{Tables}
\crefname{table}{Tab.}{Tabs.}

\begin{document}
\title{Weakly Supervised 3D Instance Segmentation \\without Instance-level Annotations}
\author{Shichao Dong, Guosheng Lin
\thanks{Shichao~Dong (e-mail: scdong@ntu.edu.sg) and Guosheng Lin are with S-Lab, Nanyang Technological University, Singapore and the School of Computer Science and Engineering, Nanyang Technological University, Singapore.}
\thanks{Corresponding author: Guosheng Lin (Email: gslin@ntu.edu.sg).}
}

\maketitle

\begin{abstract}
3D semantic scene understanding tasks have achieved great success with the emergence of deep learning, but often require a huge amount of manually annotated training data. To alleviate the annotation cost, we propose the first weakly-supervised 3D instance segmentation method that only requires categorical semantic labels as supervision, and we do not need instance-level labels. The required semantic annotations can be either dense or extreme sparse (e.g. 0.02\% of total points). Even without having any instance-related ground-truth, we design an approach to break point clouds into raw fragments and find the most confident samples for learning instance centroids. Furthermore, we construct a recomposed dataset using pseudo instances, which is used to learn our defined multilevel shape-aware objectness signal. An asymmetrical object inference algorithm is followed to process core points and boundary points with different strategies, and generate high-quality pseudo instance labels to guide iterative training. Experiments demonstrate that our method can achieve comparable results with recent fully supervised methods. By generating pseudo instance labels from categorical semantic labels, our designed approach can also assist existing methods for learning 3D instance segmentation at reduced annotation cost.
\end{abstract}


\section{Introduction}
As a fundamental perception task in scene understanding, 3D instance segmentation is to jointly estimate the class labels and object labels for point cloud data. It has a wide range of real-world applications, such as indoor robot navigation, augmented reality, autonomous driving, etc. This research topic has achieved significant progress with recent breakthroughs in deep learning techniques. However, most of these approaches require a large number of training data with dense point-level instance annotations. Even using a segment-assisted tool, the average time for a human to annotate one scene in ScanNet \cite{dai2017scannet} is about 22.3 minutes.

There are some recent attempts \cite{3d_weak_ins_box,otoc_Liu_2021_CVPR,tao2020seggroup,wang2020weakly,wei2020multi,xu2020weakly} trying to tackle the weakly supervised segmentation problem on 3D point clouds, with various weak label settings. However, some defined weak annotation methods are hard to follow in practice. For example, Xu et al. \cite{xu2020weakly} uses 10\% of total labeled points, which is sparser but uniformly distributed. This is achieved by sub-sampling, but can hardly reduce the burden of annotation in practice. Besides, the performance gap between most weakly supervised methods and their fully-supervised baselines is still large. To this end, some more effective and practical solutions are highly desirable.

\begin{figure}[t]
	\begin{center}
		\includegraphics[width=1.0\linewidth]{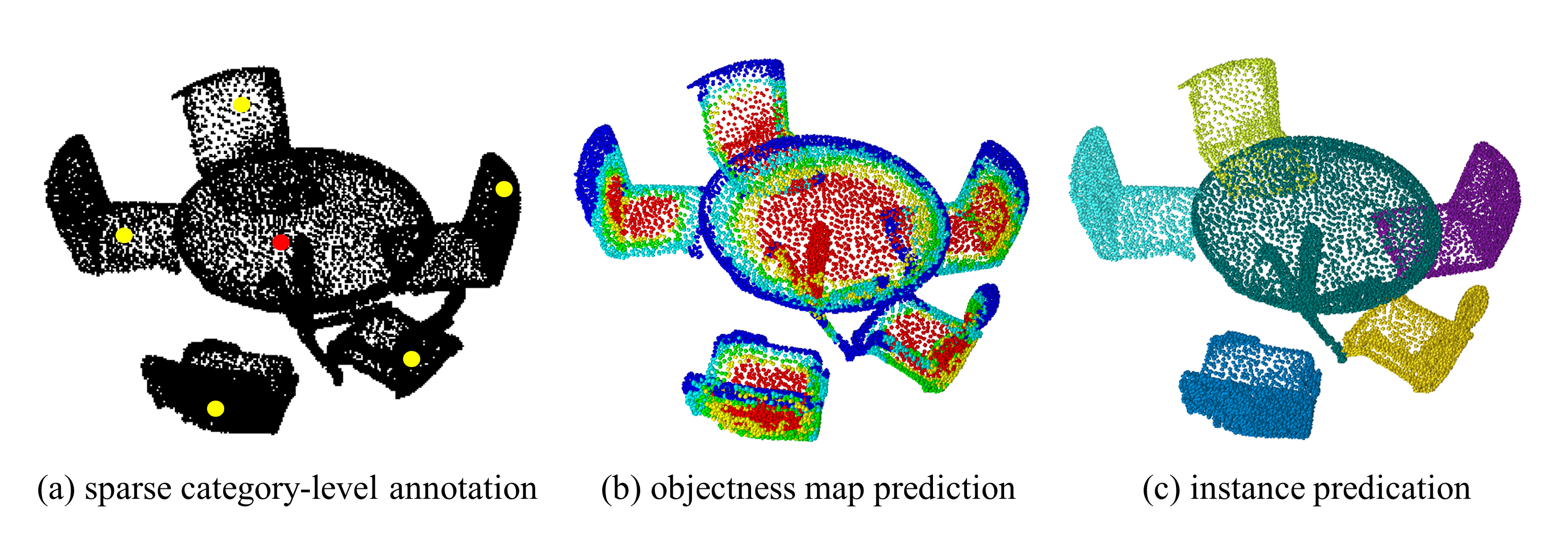}
	\end{center}
	\caption{Given an input point cloud and its category-level annotation (a) only, our task is to predict instance segmentation (c). Our method define a novel objectness signal as (b) to encode instance-aware contextual information. Note that (a) is for illustration purpose, our method support different settings for dense or random sparse semantic annotations (e.g. 0.02\%, 1\%, etc). Semantic labels can be annotated in various forms without any strict constraints.}
	\label{fig:illustration}
\end{figure}

In this paper, we introduce a novel weakly-supervised framework for 3D instance segmentation that only relies on semantic labels. Unlike semantic labels that fall into a fixed number of categories, instance labels are sometimes more tedious to prepare, since points belonging to different objects are to be labelled with nonidentical ids. For a scene with 20 chairs, semantic annotation can be done in a single shot but instance labels have to be annotated object by object, which can cost relatively more human effort and time. Our method also support very sparse annotations (i.e. 0.02$\%$), which can further reduce the annotation cost.

Existing weakly supervised 3D instance segmentation methods \cite{3d_weak_ins_box,tao2020seggroup} are still provided with rough information of localized instances. The main challenge of our task is to extract the instance-wise information from the data itself without any prior knowledge. To address this issue, we first use the void spaces between objects as clues to make an initial guess and gradually improve the ability of recognition with our designed strategy.

As illustrated in Figure \ref{fig:illustration}, we only use semantic annotations as supervision and aim to perform instance segmentation. Besides instance centroid regression signal, we also define a novel learnable objectness signal. The results show it can reflect very important instance-aware information. Further, we design a novel algorithm that can leverage the power of the objectness signal. The key contributions of this paper can be summarized as follows:

\begin{itemize}
    \item We propose a novel weakly supervised learning approach for 3D point cloud instance segmentation tasks using only categorical semantic labels (dense or extremely sparse, e.g. 0.02\%), and we do not require any instance-level labels for training. To the best of our knowledge, this is the first work to solve 3D instance segmentation task without any instance-wise annotations.
    
    \item We define a novel concept of objectness signal with multilevel shape-adaptive patterns to encode instance-aware contextual information. We also design rules to randomly generate virtual scenes with pseudo instances, enabling us to construct a recomposed dataset that enhances the learning of objectness signals.
    
    \item We propose an asymmetrical object inference algorithm that can treat points with different strategies based on their objectness prediction, which helps to resolve under-segmentation and missing detection issues.
    
    \item Even without instance-level annotations, our method achieves comparable results with fully supervised method in 3D instance segmentation.

\end{itemize}

\section{Related Work}

\subsection{Point cloud segmentation}
To deal with point cloud data \cite{bello2020review} which is sparsely distributed in space, there are generally two types of approaches. Point-based methods \cite{PointCNN,PointNet,PointNet++,KPConv,PointConv,PointWeb,9552005,Guo_2021} directly deal with the unstructured and unordered points based on their original coordinates. Whereas the other type of methods \cite{3DSIS,multiview2,Submanifold,MVPNet,multiview} transform sparsely distributed point cloud data into fixed 3D volumetric grids.

\subsection{Fully supervised 3D instance segmentation}
Most existing 3D instance segmentation methods are fully supervised, which requires point-level annotation of semantic label and instance label. As a joint task of segmentation and localization, there are two types of approaches: proposal-based methods \cite{SGPN,3DSIS,3DBoNet,3dmpa} and proposal-free methods \cite{SGPN,MASC,ASIS,JSIS3D,MTML,pointgroup,occuseg,Chen_2021_ICCV,Liang_2021_ICCV,DBLP:journals/tmm/LiuGMLW21}. The former performs an object detection task first and then predicts a mask for each proposal. The latter does not explicitly detect bounding boxes for objects, but instead it groups points on the feature embeddings. Jiang et al. \cite{pointgroup} adopt a breadth-first search clustering algorithm on dual coordinate sets. Some other recent works \cite{Chen_2021_ICCV,Liang_2021_ICCV,dong2022learning,vu2022softgroup,2022arXiv220314662Z} are built on top of PointGroup \cite{pointgroup} with further improvement.

\subsection{Weakly supervised point cloud segmentation}
Many previous weakly supervised methods \cite{ahn2018learning,huang2018weakly,pinheiro2015image,papandreou2015weakly,ahn2019weakly,zhou2018weakly,arun2020weakly,DBLP:journals/tmm/ZhangYZJL23,DBLP:journals/tmm/LiJ00W023,DBLP:journals/tmm/0012YZ0XS23,DBLP:journals/tmm/ShuL0B023,DBLP:journals/tmm/Gama0JS23,DBLP:journals/tmm/LiuKHL23,DBLP:journals/tmm/ZhouGLF21} are on 2D images. Wei et al. \cite{wei2020multi} first propose to use Class Activation Map (CAM) and attention modules for weakly supervised point cloud semantic segmentation, with subcloud-level labels. Xu et al. \cite{xu2020weakly} propose to use sparser weak labels for supervision. Seg-level weak label setting was used in \cite{otoc_Liu_2021_CVPR,tao2020seggroup}, which annotates one point for each object and then propagates to all other points within the same supervoxel as their initial pseudo label. There are only a few attempts for weakly supervised instance segmentation. As a pioneer work, Tao et al. \cite{tao2020seggroup} introduced to use graph convolution network (GCN) to propagate instance label. Liao et al. \cite{3d_weak_ins_box} and J. Chibane et al. \cite{chibane2021box2mask} suggested the use of 3D bounding boxes as a form of supervision for this task. It should be noted, however, that the annotation of 3D bounding boxes provides a more comprehensive set of information. Many non-overlapping objects can already be delineated by their respective 3D bounding boxes. All these instance segmentation approaches have to utilize instance-related ground-truth as a form of guidance or reference in order to be effective.

\begin{figure*}[t]
	\begin{center}
		\includegraphics[width=1.0\linewidth]{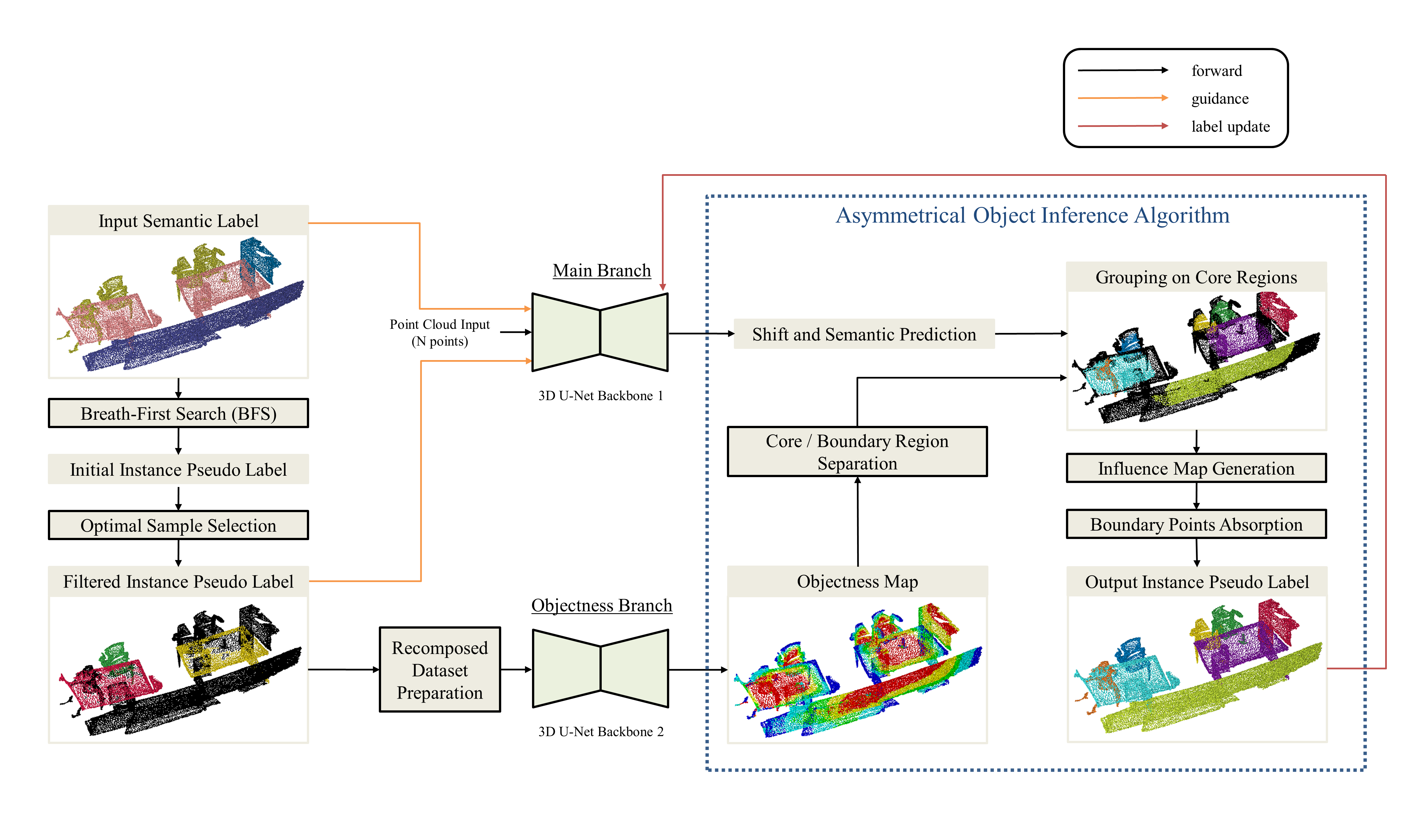} 
	\end{center}
	\vspace{-8mm}
	\caption{Overview of proposed weakly supervised method for 3D instance segmentation. Based on the categorical semantic labels on training data, we first employ a breadth-first search algorithm to find rough regions based on the geometrical connectivity of points. Then we select those confident samples for main branch and objectness branch. Submanifold sparse convolution \cite{Submanifold} based 3D U-Net backbones are used to extract point features. The objectness backbone is trained on a recomposed dataset for predicting real data. Our proposed Asymmetrical Object Inference Algorithm (AOIA) considers the input signals from both branches and generates instance pseudo labels to supervise further training. Note that input semantic labels can be either from ground-truth labels or generated pseudo labels.}
	\label{fig:Pipeline}
\end{figure*}

\section{Methods}

\subsection{Architecture Overview}
\label{sec:ntwk}
In Figure \ref{fig:Pipeline}, the pipeline of our approach can be divided into three steps: sample preparation, feature learning, and object inference. The first step is to find confident samples and reduce the noise impact. Then, our network architecture consists of two separated networks for learning useful instance-aware features. In the last step, we propose an object inference algorithm to group points with the guidance of the predicted signals from the previous stage. We use this algorithm for generating high-quality instance pseudo labels on the training set, as well as making predictions on unseen data. 

\subsection{Initial Instance Pseudo Label Generation}
\label{sec:bfs}
In this work, the predominant objective is to discover the instance cues from the data itself, because there is no human input at all. Unlike 2D images that lack of depth information, objects in 3D point clouds are usually well kept as their source distribution and naturally separated. Most instances are not closely placed. Even without any sense of instance at this stage, we can utilize the void space between distanced objects and perform a rough grouping based on their original coordinates.

As shown in the left part of Figure \ref{fig:Pipeline} and Algorithm \ref{algo:algorithm_bfs}, we first employ a widely used breadth-first search algorithm \cite{pointgroup} to find connected regions for foreground categories. For input point cloud $\bm P \in \mathbb{R}^{N \times 3}$, a typical step of BFS is about searching all points within the radius $r$ from seed point $P_i$ in Euclidean space. If a target point has the same semantic label as the seed point, it is then grouped into the same cluster. This simple operation repeats on each of the newly grouped points until no further points can be grouped. This straightforward algorithm suits well in point cloud data and helps to find those geometrically connected regions.

\begin{algorithm*}
	\caption{Breadth-First Search (BFS) Algorithm}
	\label{alg_cluster}
	\hspace*{0.03in} {\bf Input:} clustering radius $r$; \\
	\hspace*{0.4in}
	cluster point number threshold $N_{\theta}$; \\
	\hspace*{0.4in}
	coordinates $\mathbf{X} = \{x_1, x_2, ..., x_N\}\in \mathbb{R}^{N \times 3}$ \\
	\hspace*{0.4in}
	semantic labels $\mathbf{S} = \{s_1, ..., s_N\}\in \mathbb{R}^{N}$. \\
	\hspace*{0.02in} {\bf Output:} 
	clusters $\mathbf{C} = \{C_1, ..., C_K\}$.
	\begin{algorithmic}[1]
		\State initialize an array $v$ (visited) of length $N$ with all zeros
		\State set all background points as visited
		\State initialize an empty cluster set $\mathbf{C}$
		\For{$i = 1$ to $ N$}
		\If{$v_i == 0$} \Comment{start from an unvisited point}
		\State initialize an empty queue $Q$
		\State initialize an empty cluster $C$
		\State $v_i = 1$; $Q$.enqueue($i$); add $i$ to $C$ \Comment{create a new cluster from a seed point}
		\EndIf
		\While{$Q$ is not empty} \Comment{iterative grouping} 
		\State $k = Q$.dequeue()
		
		\For{$j \in [1, N]$ with $||x_j - x_k||_2 < r$} \Comment{find target points within distance of $r$}
		\If{$s_j == s_k$ and $v_j == 0$} \Comment{group satisfied target points}
		\State $v_j = 1$; $Q$.enqueue($j$);add $j$ to $C$
		\EndIf
		\EndFor
		\EndWhile
		
		\If{$C > $$N_{\theta}$} \Comment{output cluster if point number in it is above threshold}
		\State add $C$ to $\mathbf{C}$
		\EndIf
		\EndFor
		
		\State \Return $\mathbf{C}$
	\end{algorithmic}
	\label{algo:algorithm_bfs}
\end{algorithm*}

\subsection{Optimal Sample Selection}
\label{sec:optimal_sample}
However, the generated initial instance pseudo labels are still not accurate enough, especially for the following two cases. First, closely placed objects from the same category can be mistakenly grouped if their minimum void distances are less than the searching radius of BFS algorithm. Second, point cloud objects also suffer from occlusion and noise. The inconsistent surface problem may cause objects to be over-clustered into multiple fragments. These samples are considered as noisy data, which have negative impact on our network training.

To get rid of these issues, we use an optimal sample selection strategy to filter out low confidence samples. For each semantic category, we sort the list of all predicted samples in ascending order by their number of points. Based on a percentage range, too large or too small samples are to be rejected and ignored for later usage. This operation keeps 40\% high confident samples and avoids the network being confused by the predictions with low confidence.

\subsection{Instance Centroid Offset Prediction}
\label{sec:offset_branch}
Learning point-wise offset vectors and shifting points to their respective instance centroids is an effective way to deal with over-segmentation and under-segmentation problems during clustering. We use submanifold sparse convolution \cite{Submanifold} based U-Net backbone network to extract point-wise feature $\bm F \in \mathbb{R}^{N \times dim}$. A two-layered MLP is followed to transform point feature $\bm F$ into centroid shift vector $\bm D \in \mathbb{R}^{N \times 3}$. For the first iteration, only selected high confidence samples from the previous step are used for supervision.

The instance centroid $\widehat{q}$ is defined as the mean coordinates of all points with the same instance label. Following previous work \cite{pointgroup}, we use L1 regression loss to train offset prediction,

\begin{multline}
    L_{offset} = \frac{1}{\sum_i \mathcal{M}_i}\sum_i ||d_i - (\hat{q}_i - p_i)|| \cdot \mathcal{M}_i - \\ \frac{1}{\sum_i \mathcal{M}_i} \sum_i \frac{d_i}{||d_i||_2} \cdot \frac{\hat{q}_i - p_i}{||\hat{q}_i - p_i||_2} \cdot \mathcal{M}_i.
\end{multline}

where $\mathcal{M} \in \{0,1\}$ is a binary mask indicating whether points belong to selected samples, $d$ is the predicted shift vector, $\hat{q}_i - p_i$ is the ground-truth shift vector. The second term in the equation is an additional direction loss based on cosine similarity.

\subsection{Semantic Prediction}
Semantic prediction is only required during evaluation on test or validation set, but not for the training set pseudo label generation. The semantic score $y$ is linearly transformed from the same point feature $f$. The training process is supervised by a conventional cross-entropy loss $H_{CE}$. We define the semantic loss as

\begin{equation}\label{eqn:ce}
	L_{semantic} = - \frac{1}{N} \sum_{i=1}^{N} H_{CE}(y_i, c_i).
\end{equation}

where $c_i$ represents the semantic label.

The joint loss function for this backbone can be written as
\begin{equation}
	L_{joint} = L_{offset} + L_{semantic}.
\end{equation}

\paragraph{Supervoxel smoothness}
\label{sec:supervoxel_smoothness}
To enhance the local consistency, we further use precomputed supervoxels to refine the semantic predictions. For each supervoxel set $\mathcal{V}=\{p_1, p_2, ...,p_i\}$, average pooling is applied on all semantic features for smoothing.

\subsection{Two Staged Learning Strategy}
Network training of the main backbone in our method consists of two stages. In the first stage, training is partially supervised by the selected initial instance pseudo labels. Afterward, higher quality labels generated from our asymmetric object inference algorithm are used to supervise the next stage of training. With this learning strategy, the quality of the learned offset feature can be further improved.

\subsection{Objectness Branch}
\label{sec:objectness_branch}
Although instance centroid offset signals can help to better separate objects, directly regressing to object centroids from every point is not an easy task in practice. The predictions may not be always accurate, especially for points near object boundary.  To solve this issue, we propose a novel approach for learning objectness, which reflects instance activation levels.

\subsubsection{Sample Extraction}
For each selected high confidence sample from previous steps, we first shift the object from its center (mean coordinates) to the origin of the coordinate system as $[0,0,0]$.

\subsubsection{Objectness Label Generation}
Assume that the input instance sample contains $\mathcal{N}$ points. Given the coordinates $\mathcal{X} = \{x_1, x_2, ..., x_{\mathcal{N}}\}\in \mathbb{R}^{\mathcal{N} \times 3}$, we first shrink all points toward its center at $[0,0,0]$ by dividing their original coordinates by $2$. We consider this compressed point cloud as a new object $\mathcal{X'} = \{x_1/2, x_2/2, ..., x_{\mathcal{N}}/2\}\in \mathbb{R}^{\mathcal{N} \times 3}$. Then, we find the nearest distance from each point in $\mathcal{X}$ to the compressed point cloud $\mathcal{X'}$. Following that, the nearest distance list $\mathcal{D} = \{d_1, d_2, ..., d_{\mathcal{N}}\}\in \mathbb{R}^{\mathcal{N} \times 1}$ is sorted in ascending order. We define following rules for generating objectness label,

\begin{equation}
    \begin{array}{l}
    \text{objectness} \\ 
    \text{label ID}
    \end{array}
	=
	\begin{cases}
		4     &     \textrm{if within } 0\%-20\% \textrm{ of list } \mathcal{D}\\
		3     &     \textrm{if within } 20\%-40\% \textrm{ of list } \mathcal{D}\\
		2     &     \textrm{if within } 40\%-60\% \textrm{ of list } \mathcal{D}\\
		1     &     \textrm{if within } 60\%-80\% \textrm{ of list } \mathcal{D}\\
		0     &     \textrm{if within } 80\%-100\% \textrm{ of list } \mathcal{D}\\
	\end{cases},
\end{equation}

\begin{figure}[t]
	\begin{center}
		\includegraphics[width=0.8\linewidth]{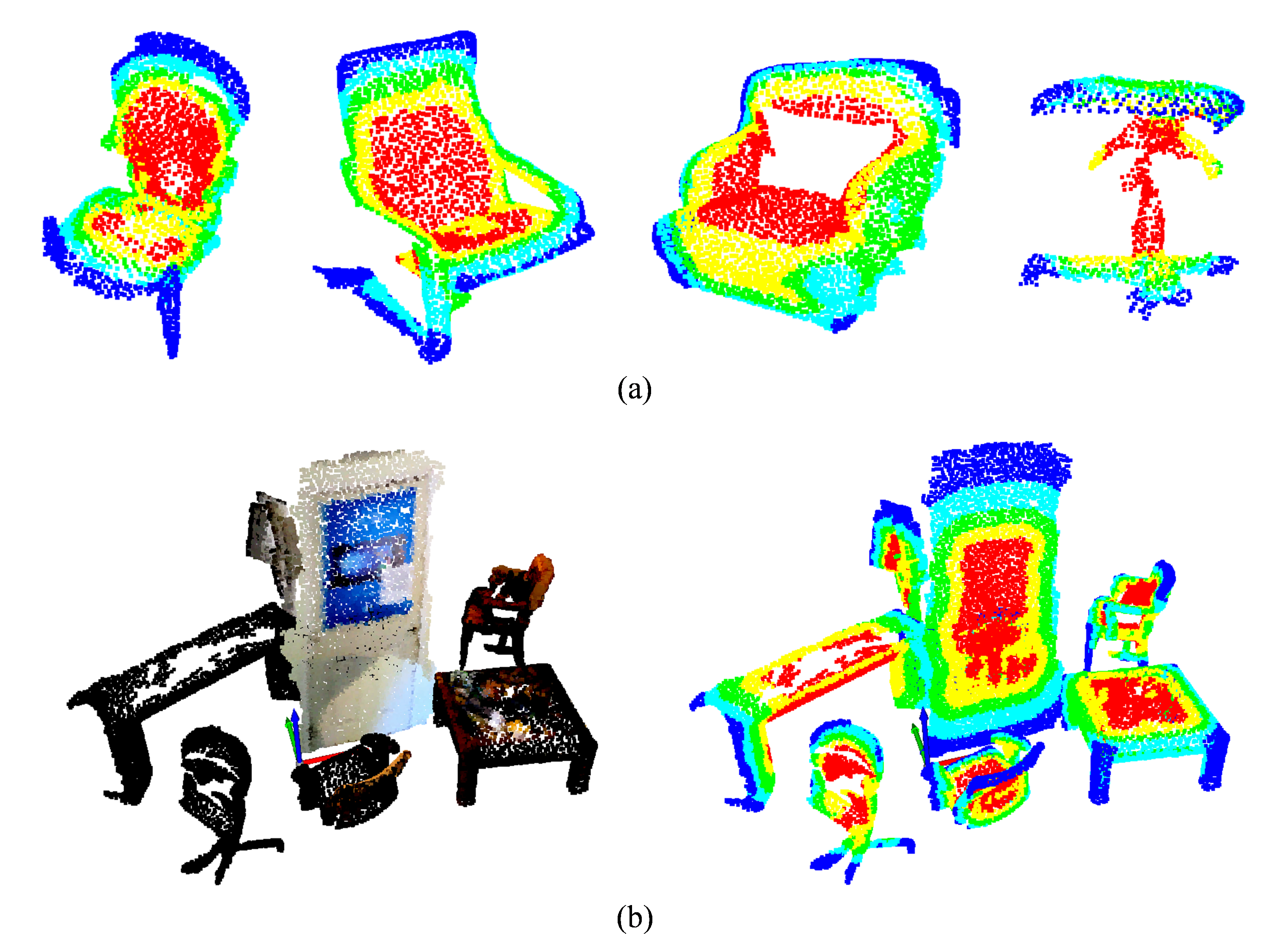}
	\end{center}
	\vspace{-4mm}
	\caption{(a) Examples of extracted instance samples with defined multilevel shape-aware objectness label. (b) An example of our randomly generated scene in the recomposed dataset.}
	\label{fig:Toydata}
\end{figure}

As shown in Figure \ref{fig:Toydata} (a), generated multilevel objectness labels are visualized in different colors (ID 4 is in red, ID 3 is in yellow, ID 2 is in green, ID 1 is in cyan, ID 0 is in dark blue ). One of the major advantages of our defined objectness map is the shape-aware character. No matter how the shape varies, our defined rule always accurately finds the inner regions of objects in an adaptive manner.

\subsubsection{Random Recomposed Dataset}
To facilitate the learning of the defined objectness signal, we create a novel dataset by recomposing selected pseudo samples. Specifically, for each virtual scene, we randomly select pseudo instance samples and then stitch them together into a flexible template. To enhance the robustness of our model, we introduce various perturbations such as random rotation, random dropping, and distance-controlled squeeze during the generation process. These perturbations aim to prevent over-fitting issues and enable the network to distinguish different instances, even when they are closely located. By adopting this data generation strategy, we obtain an unlimited and cost-effective source of data, which helps to improve the performance of our network.

\subsubsection{Learning Objectness Signal}
We use a separated U-Net backbone for learning the objectness signal on the recomposed dataset. The objectness score $q$ is linearly transformed from the point feature. The training process is supervised by a conventional cross-entropy loss $H_{CE}$. We define the objectness loss as

\begin{equation}\label{eqn:ce}
	L_{objectness} = - \frac{1}{N} \sum_{i=1}^{N} H_{CE}(q_i, o_i).
\end{equation}

where $o_i$ represents the objectness label.

\subsection{Asymmetric Object Inference Algorithm}
\label{sec:AOIA}

In this section, we introduce an asymmetric object inference algorithm to tackle the 3D instance segmentation task, based on the predicted features from the previous stage. Unlike most inference algorithms, points are not equally treated in our algorithm. Specifically, we distinguish points according to the predicted objectness map and uses different grouping strategies for high objectness regions and boundary regions. Notably, information only propagates from high objectness points to low objectness points, not the other way around.

The core regions of an object are the inner parts that have high objectness scores. We hypothesize that the instance centroid offset predictions for these core regions are more reliable and trustworthy. On the other hand, points located in the boundary regions of an object often pose a challenge in predicting the precise centroid offset. If the shift prediction is inaccurate, this can result in nearby objects being mistakenly merged together.

\begin{figure}[t]
	\begin{center}
		\includegraphics[width=1.0\linewidth]{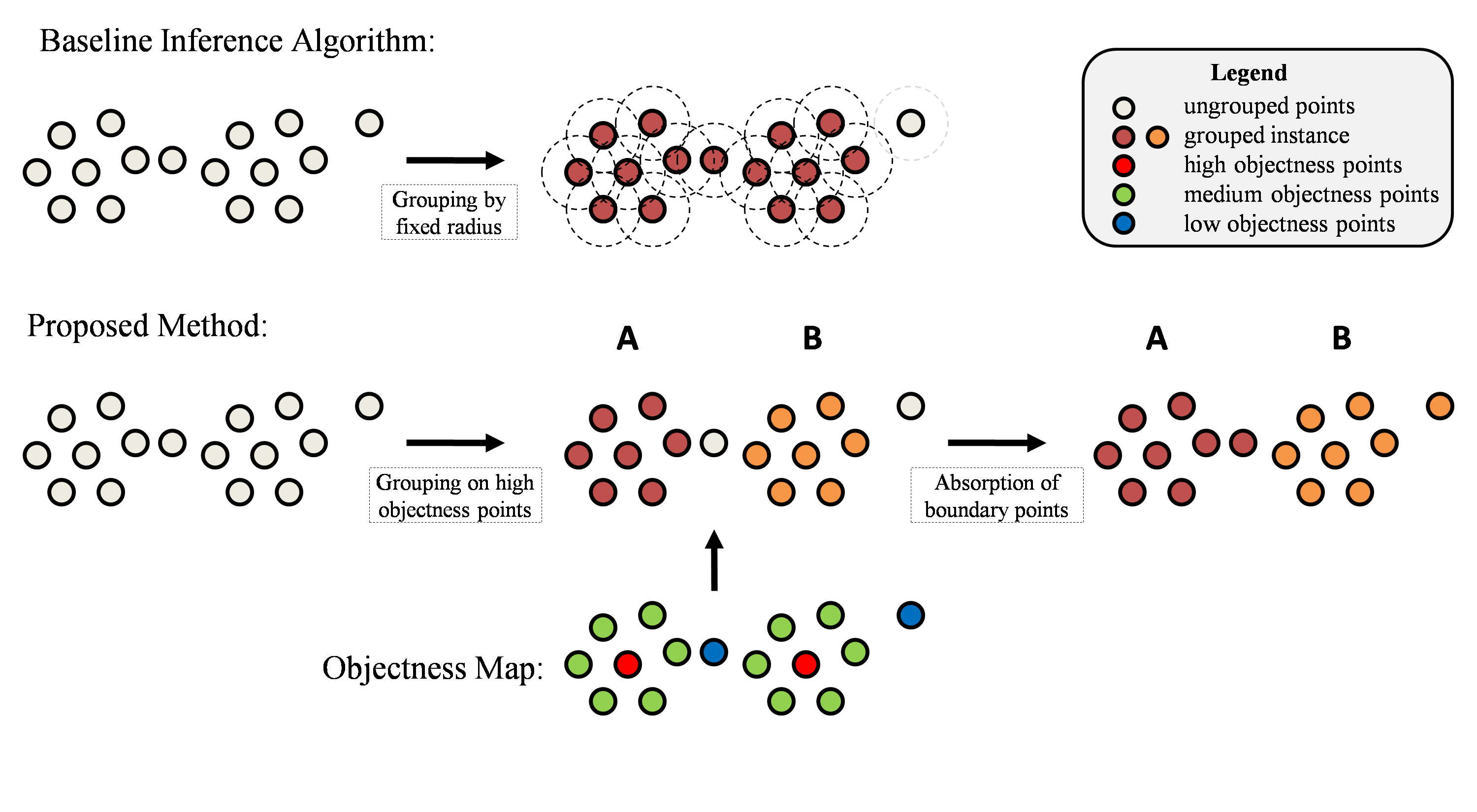}
	\end{center}
	\caption{Illustration of proposed asymmetric object inference algorithm.}
	\label{fig:AOIA}
\end{figure}

In Figure \ref{fig:AOIA}, we compare our proposed method with the baseline inference algorithm. The baseline method groups input points based on their geometrical connectivity, whereby points are grouped together if the euclidean distance between them is less than a fixed radius $r$. However, the point located between two clusters served as a bridge and caused two objects to be mistakenly grouped into one, while the right-most point was also missed from being grouped.

In our implementation, we initially ignore the low objectness points (in blue) to prevent any potential negative impact. Then,we apply the searching process on the core points, which can lead to the formation of two separate clusters, represented as $\bm A$ and $\bm B$. Lastly, we employ an attraction-based method to further absorb any ungrouped boundary points. Our method offers two major advantages compared to the baseline: (1) it helps avoid under-segmentation issues and (2) it provides better grouping of discrete points.

\begin{figure}[thb]
	\begin{center}
		\includegraphics[width=0.5\linewidth]{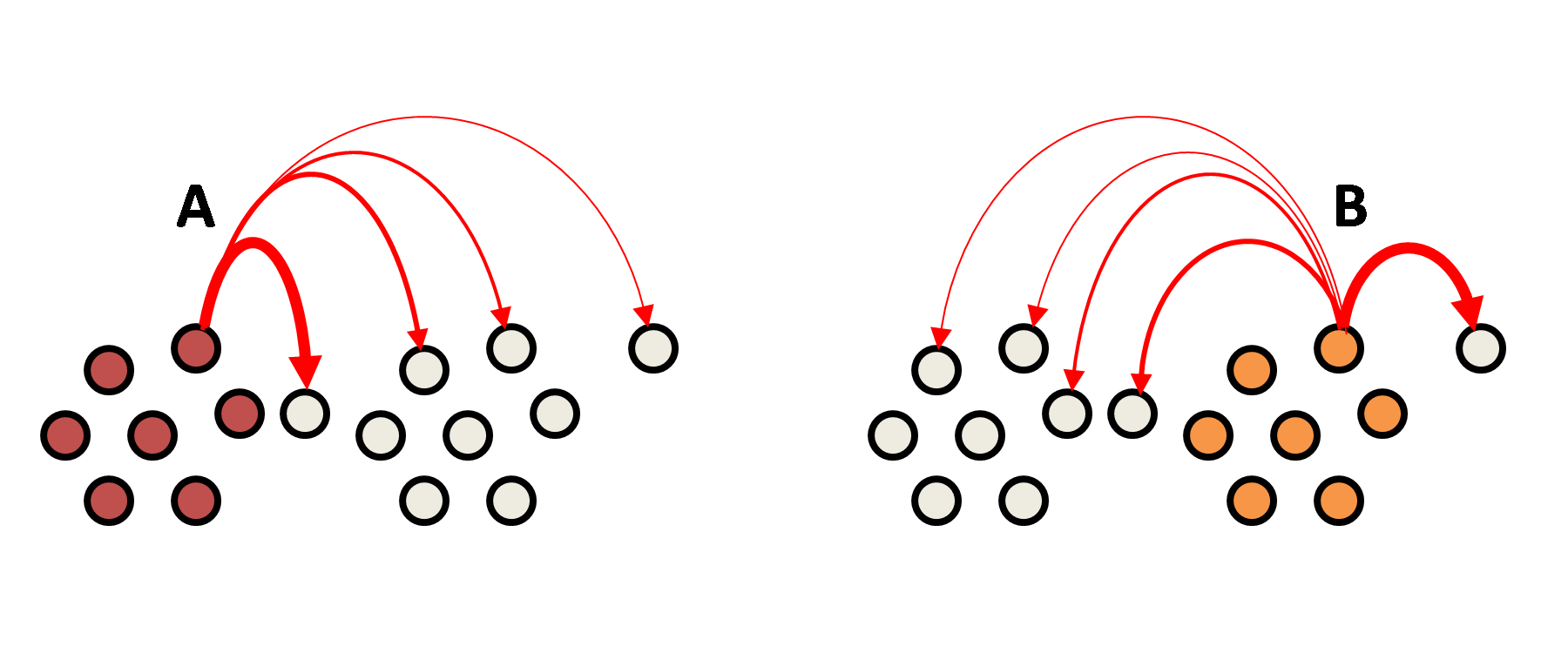}
	\end{center}
	\caption{Illustration of influence maps. We define high objectness points to have the ability to influence neighbouring points. The pairwise influence depends on the initial potential value $v_i$ and their relative distances based weights $w$ (shown as the thickness of red arrow lines). The total influence from a core cluster to a target low objectness point equals to the sum of all pairwise influence. A low objectness point is to be absorbed to a core cluster that has highest influence on it.}
	\label{fig:influence_map}
\end{figure}

To facilitate the absorption step, we compute an influence map on intra-category points for each core point cluster. In particular, the fully connected pairwise influence weight from point $p_i$ to point $p_j$ $(p_i, p_j \in \mathcal{P}_{fg})$ can be expressed as

\begin{equation}\label{eq:W}
	\bm w_{ij} = 
	\begin{cases}
		 $0$ & (i,j) \in \mathcal{P}_{core},\\
		{\rm \exp}  (- \frac{\| x_i - x_j \|^2}{2\varepsilon^2}) & \textrm{otherwise}
	\end{cases},
\end{equation}

where $x$ is the coordinates of points, $\varepsilon$ is a hyperparameter. $\mathcal{P}_{fg}$ represents points from the same foreground semantic category. $\mathcal{P}_{core}$ represents points from the same core cluster.

As shown in Figure \ref{fig:influence_map}, the value of $w$ depends on the distance between two points. Closer points have higher influence weights. To balance the effect from different clusters, we define the potential values on core point cluster $\bm A$ as

\begin{equation}\label{eq:v}
    {\bm v}_{i}^{A} = \frac{1}{|\mathcal{P}_A|}
\end{equation}

For a target point $j$, the total influence from core cluster $\bm A$ is defined as

\begin{equation}\label{eq:I}
    \mathcal{I}_{j}^{A} = \sum_{i \in \mathcal{P}_A} {\bm v}_{i}^{A} \times \bm w_{ij}
\end{equation}

Therefore, we have $\mathcal{I}_{j} = \{\mathcal{I}_{j}^{A}, \mathcal{I}_{j}^{B}, ..., \mathcal{I}_{j}^{K}\}\in \mathbb{R}^{K \times 1}$, where $K$ is the number of core clusters. Last, we group boundary point $j$ into cluster $N$, if $\mathcal{I}_{j}^{N}$ has the highest influence level among all core clusters. The complete inference procedure is detailed in Algorithm \ref{algo:algorithm_AOIA}.

\begin{algorithm}[t]
    \caption{\small Asymmetric Object Inference Algorithm}
	\hspace*{0.02in} {\bf Input:} coordinates $\mathbf{X}\in \mathbb{R}^{N \times 3}$; semantic label $\bm S \in \mathbb{R}^{N \times 1}$; offset prediction $\bm D \in \mathbb{R}^{N \times 3}$; objectness map $\bm O \in \mathbb{R}^{N \times 1}$, objectness id threshold $\hat{O}$\\
	\hspace*{0.02in}{\bf Output:} 
	Instance pseudo label $\bm C$ \\
	\begin{algorithmic}[1]
	    \For{$id$ $\in$ $unique(\bm S)$}
	    \If{$id$ $\in$ foreground semantic IDs}
	    \State ${N}_{fg}  \gets |\mathcal{P}_{fg}| $
	    
	    \For{$i$ = 1 to $N_{fg}$}
	    \If{$\bm O_{fg} >= \hat{O}$}
	    \State $\mathcal{P}_{core}  \gets \mathcal{P}_{i} $
	    \Else
	    \State $\mathcal{P}_{boundary}  \gets \mathcal{P}_{i} $ 
	    \EndIf
	    \EndFor
        
        \Comment{shift points to predicted centers}
        \State $\mathbf{X'}_{fg} = \mathbf{X}_{fg} + \mathbf{D}_{fg} $
	    
	    \Comment{group core regions into $K$ clusters}
	    \State ${C}_{core}^{K}  \gets \mathcal{P}_{core} $
	    
	    \For{$k$ = 1 to $K$}
	    \State compute influence map $\mathcal{I}^{k}$
	    \State $\mathcal{I} \gets \mathcal{I}^{k}$
	    \EndFor
	    
	    \For{$\mathcal{P}_{j} \in \mathcal{P}_{boundary}$}
	    \State compare influence levels among $\mathcal{I}_{j}$
	    \State $k \gets \argmax{(\mathcal{I}_{j})}$
	    
	    \Comment{group boundary points into an cluster}
	    \State ${C}_{core}^{k} \gets \mathcal{P}_{j} $
	    \EndFor
	    \State ${C}_{id} \gets {C}_{core}$
	    \EndIf
	    \EndFor
		\State \Return $\mathbf{C}$
		
	\end{algorithmic}
	\label{algo:algorithm_AOIA}
\end{algorithm}

\section{Experiment}
\label{sec:exp}
We evaluate the proposed method on two challenging 3D indoor scene public datasets: ScanNet-v2 \cite{dai2017scannet} and S3DIS \cite{S3DIS}. We first evaluate the quality of the generated pseudo labels, in comparison with the target ground-truth labels on the training set of ScanNet-v2 \cite{dai2017scannet}. Then, we compare the prediction performance of our method with other existing weakly supervised and fully supervised methods. To demonstrate the effectiveness of each component in the proposed method, we further conduct comprehensive ablation studies. 

\subsection{Implementation details}
We have two separate backbone networks for the main branch and the objectness branch. Their submanifold sparse convolution \cite{Submanifold} based U-Net structure are identical. Following the same backbone parameters in \cite{pointgroup,Chen_2021_ICCV}, we use the voxel size of $2 cm$ and $7$ layers of U-Net. The batch size is set as $8$ and $16$ respectively for two backbones. The main backbone network is trained for two stages and the objectness backbone network is just trained in a single stage on $5000$ virtual scenes in the recomposed dataset. The whole training process is on a single NVIDIA RTX 3090-ti GPU card, using Adam solver for optimization and an initial learning rate of $0.001$. 

For asymmetric object inference algorithm, the hyperparameter $\varepsilon$ of radial basis function kernel in Equation \ref{eq:W} is set to $3$. For final prediction, the BFS grouping radius for core points in Algorithm \ref{algo:algorithm_AOIA} is set to $5cm$, and the objectness id threshold $\hat{O}$ is set to $1$, which means considering top $80\%$ as core points. The computation of influence maps is accelerated by GPU. We subsample the number of the input point $\mathcal{P}_{fg}$ by the voxel size of $5cm$ if it is above $20k$.

\paragraph{Solutions for Sparse Semantic Annotations}
Proposed method supports both dense and sparse semantic annotations. For sparse annotations, we have evaluated two options: ``XMFR'' and OTOC\cite{otoc_Liu_2021_CVPR}. These methods can generate pseudo semantic labels from very sparse annotations (i.e. $0.02\%$). This can be further utilized by our main pipeline for dense prediction of instance segmentation.

\begin{figure*}
	\begin{center}
		\includegraphics[width=0.99\linewidth]{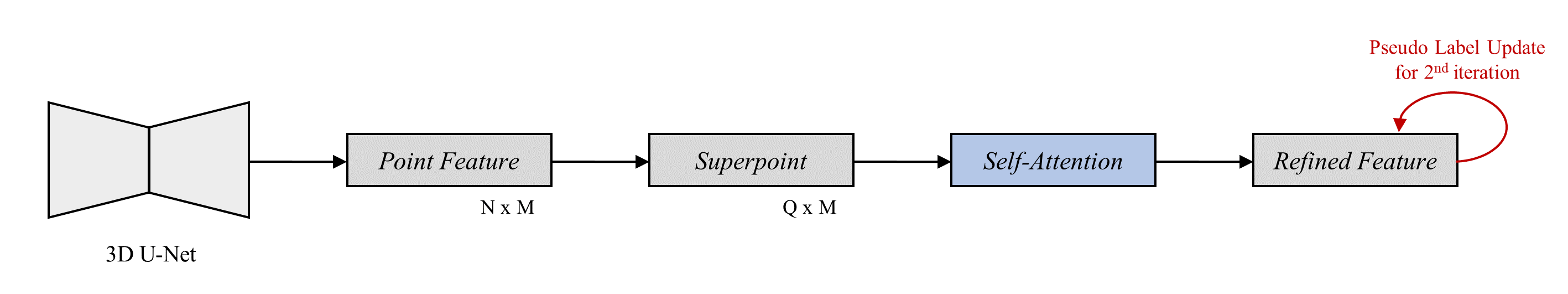}
	\end{center}
	\caption{Pipeline of proposed self-attention based method. $N$ is the number of point, $M$ is the feature dimension, $Q$ is the number of superpoint. Generated pseudo labels are further used to supervise the second round training.}
	\label{fig:supp_figure7}
\end{figure*}

We propose a simple self-attention based method for semantic label propagation. As shown in Figure \ref{fig:supp_figure7}, it uses a same submanifold sparse convolution \cite{3DSemanticSegmentationWithSubmanifoldSparseConvNet} based 3D U-Net backbone for feature extraction. Extracted point features are further grouped into superpoints, based on the generated supervoxels by \cite{Lin2018Supervoxel}. Average pooling is applied for each supervoxel to have unique features. Following \cite{zhao2021pointtransformer}, we use a point transformer layer for semantic feature propagation. The major advantage is that it has a global receptive field and helps to propagate information to those unknown regions. Refined features are interpolated to the original point resolution. The training process is supervised by a conventional cross-entropy loss. Both training and evaluation is performed on the training set data. For the first round training, we spread the sparse semantic labels to their neighbouring points under the same supervoxels as ground-truth. The produced dense pseudo semantic labels are further used to supervise the second round training, which can further boost the performance.

Besides this self-attention based method, our pipeline is compatible with all existing weakly supervised semantic segmentation methods, such as \cite{otoc_Liu_2021_CVPR,wei2020multi}. Without any additionally required annotations, our work can extend these weakly supervised semantic segmentation methods for an instance segmentation task at no extra cost.

\subsection{ScanNet-v2 Dataset}
ScanNet-v2 dataset \cite{dai2017scannet} is a well-recognized 3D point cloud dataset for indoor scenes. It contains 2.5 million real-world RGB-D views and is reconstructed into 1513 indoor scenes. The 3D meshed data are annotated with semantic labels and instance labels for 20 categories.

\begin{table*} [thb]
    \resizebox{\textwidth}{!}{
    \begin{tabular}{c|cccccccccccccccccc|c}
    \toprule
    \textbf{Metric} & cab & bed & chair & sofa & tabl & door & wind & bkshf & pic & cntr & desk & curt & fridg & showr & toil & sink & bath & ofurn & \textbf{avg}\\
    \midrule
    mAP&54.6&88.4&72.5&88.2&80.8&52.6&72.6&70.4&76.0&69.6&61.6&62.8&94.1&89.0&99.0&92.3&96.5&63.3&\textbf{76.9}\\
    mAP@0.5&65.9&90.3&81.9&92.3&86.2&69.6&82.9&80.8&82.4&81.7&72.8&79.0&96.0&93.5&99.3&95.9&97.5&73.0&\textbf{84.5}\\
    mAP@0.25&73.1&90.3&84.8&93.8&87.0&73.4&86.0&82.7&84.9&84.7&77.2&79.9&96.0&93.5&99.3&96.2&97.5&78.2&\textbf{86.6}\\
    \bottomrule
    \end{tabular}
    }
    \caption{Instance pseudo label quality on ScanNet v2~\cite{dai2017scannet} training set.}
    \label{tab:instance_segmentation_scannet_train}
\end{table*}

\begin{table*}
    \resizebox{\textwidth}{!}{
    \begin{tabular}{c|cccccccccccccccccc|c}
    \toprule
    \textbf{Metric} & cab & bed & chair & sofa & tabl & door & wind & bkshf & pic & cntr & desk & curt & fridg & showr & toil & sink & bath & ofurn & \textbf{avg}\\
    \midrule
    mAP&    25.6&42.0&59.7&42.6&34.9&26.1&28.6&33.3&35.4&11.7&16.8&25.0&38.6&53.0&78.5&45.9&55.2&32.9&\textbf{38.1}\\
    mAP@0.5&44.3&70.5&74.4&67.2&54.8&41.9&47.5&56.6&47.7&32.6&39.4&44.1&49.5&73.1&95.8&65.8&74.0&47.6&\textbf{57.0}\\
    mAP@0.25&62.5&82.3&82.2&79.3&69.6&57.5&62.3&71.0&54.2&65.5&72.4&55.7&52.7&81.9&98.3&84.5&79.2&60.3&\textbf{70.6}\\
    \bottomrule
    \end{tabular}
    }
    \caption{Instance segmentation results on ScanNet v2~\cite{dai2017scannet} validation set.}
    \label{tab:instance_segmentation_scannet_val}
\end{table*}

\subsubsection{Evaluations of instance pseudo label quality}
In Table \ref{tab:instance_segmentation_scannet_train}, we present the evaluation results of our generated pseudo label on full training set, with 1201 scenes. Reported pseudo labels are generated by our asymmetric object inference algorithm based on the offset prediction after two stages of training on the main backbone network and the objectness prediction after the first stage. Since no previous methods have published their instance pseudo label results, we can only provide our results without comparisons. However, the obtained result values themselves indicate a satisfactory quality of our method.

\begin{table}[b]
	\footnotesize
	\begin{center}
	    \scalebox{0.8}{
		\begin{tabular}{c|c|c|ccc}
			\toprule
			Method  & Sem. Sprv. & Ins. Sprv.  & \hspace{3pt} AP & \hspace{3pt}AP{\tiny 50}\hspace{3pt} & AP{\tiny 25}\\
			\midrule
			PointGroup~\cite{pointgroup} &100\% & 100\% & 34.8 & \hspace{-3pt}56.9\hspace{-3pt} & \textbf{71.3}\\
			\midrule
			CSC-20 (PointGroup)~\cite{hou2021exploring} & 20 pts & 20 pts & - & \hspace{-3pt}27.2\hspace{-3pt} & - \\
			CSC-50 (PointGroup)~\cite{hou2021exploring} & 50 pts & 50 pts & - & \hspace{-3pt}35.7\hspace{-3pt} & - \\
			CSC-100 (PointGroup)~\cite{hou2021exploring} & 100 pts & 100 pts & - & \hspace{-3pt}43.6\hspace{-3pt} & - \\
			CSC-200 (PointGroup)~\cite{hou2021exploring} & 200 pts & 200 pts & - & \hspace{-3pt}50.4\hspace{-3pt} & - \\
			\midrule
            TWIST \cite{TWIST9879061} & 1\% & 1\% & 9.6 & \hspace{-3pt}17.1\hspace{-3pt} & 26.2 \\
			TWIST \cite{TWIST9879061} & 5\% & 5\% & 27.0 & \hspace{-3pt}44.1\hspace{-3pt} & 56.2 \\
			TWIST \cite{TWIST9879061} & 10\% & 10\% & 30.6 & \hspace{-3pt}49.7\hspace{-3pt} & 63.0 \\
			TWIST \cite{TWIST9879061} & 20\% & 20\% & 32.8 & \hspace{-3pt}52.9\hspace{-3pt} & 66.8 \\
			\midrule
			SegGroup (PointGroup) \cite{tao2020seggroup} &0.02\% &0.02\% & ~23.4 & \hspace{-3pt}43.4\hspace{-3pt} & 62.9 \\
            3D-WSIS \cite{tang20223dwsis} & 0.02\% & 0.02\% & 28.1 & \hspace{-3pt}47.2\hspace{-3pt} & 67.5 \\
			RWSeg \cite{RWSeg} &0.02\% &0.02\% & ~34.7 & \hspace{-3pt} 56.4\hspace{-3pt} & 71.2 \\
			\midrule
			SPIB \cite{3d_weak_ins_box} & 100\% Box & 100\% Box & - & \hspace{-3pt}38.6\hspace{-3pt} & 61.4 \\
            Box2Mask \cite{chibane2021box2mask} & 100\% Box & 100\% Box & - & \hspace{-3pt}59.7\hspace{-3pt} & 71.8 \\
			\midrule
			\textbf{ AOIA (Ours) } &100\% &0\% & ~33.7 & \hspace{-3pt}54.9\hspace{-3pt} & 70.4 \\
			\textbf{ AOIA$^\dag$ (Ours) } &100\% &0\% & ~\textbf{38.1} & \hspace{-3pt}\textbf{57.0}\hspace{-3pt} & 70.6 \\
			\textbf{ AOIA$^\dag$ (Ours) + XMFR$^\ddag$ } &0.02\% &0\% & ~34.5 & \hspace{-3pt}52.0\hspace{-3pt} & 68.5 \\
			\textbf{ AOIA$^\dag$ (Ours) + OTOC$^\ddag$ } &0.02\% &0\% & ~35.4 & \hspace{-3pt}52.6\hspace{-3pt} & 68.9 \\
			\bottomrule
		\end{tabular}
		}
	\end{center}
	\caption{Instance segmentation results on ScanNet-v2 \cite{dai2017scannet} validation set. Methods marked with brackets in the names represents using generated pseudo labels to train another fully-supervised method (such as PointGroup) for evaluation. $^\dag$ represents using refined semantic prediction via supervoxel smoothing, as introduced in \ref{sec:supervoxel_smoothness}. $^\ddag$ represents using a pseudo semantic label generation method for sparse annotations.}
	\label{tab:instance_segmentation_scannet_val_compare}
\end{table}

Our proposed methodology produces high-quality instance pseudo labels that are capable of supervising any fully supervised method. The use of these labels significantly reduces the need for manual annotation, thereby reducing the overall annotation cost. Our method focuses solely on the input data and does not require any additional data or pretraining assistance. It can be treated as an effective automatic instance label generation tool, which could benefit both research and practice. 

\subsubsection{Prediction evaluation}
In Table \ref{tab:instance_segmentation_scannet_val}, we present the detailed evaluation results of our instance segmentation prediction on the validation set of ScanNet-v2 dataset \cite{dai2017scannet} in 18 foreground categories. In Table \ref{tab:instance_segmentation_scannet_val_compare}, we compare our method with other existing methods for instance segmentation prediction. These methods use different labeling strategy, thus the comparison is for reference purpose only. PointGroup \cite{pointgroup} is an important fully supervised method that is widely used for comparison. CSC \cite{hou2021exploring} uses a pretrained model as initialization and selects numbers of most representative points for a human to active label. Weakly supervised method SegGroup \cite{tao2020seggroup} and RWSeg \cite{dong2022learning} labels every instance by a single point and then spread the labels. The total labeled points are around 0.02\% by statistics. SPIB \cite{3d_weak_ins_box} and Box2Mask \cite{chibane2021box2mask} use boundary boxes as supervision. Unlike all existing methods, our method only uses semantic supervision and achieves comparable results with the fully supervised method \cite{pointgroup}. Our method is capable of accommodating various types of ground-truth semantic labels, including dense annotation, randomly sparse annotation, etc. 

\begin{figure}[t]
	\begin{center}
		\includegraphics[width=1.0\linewidth]{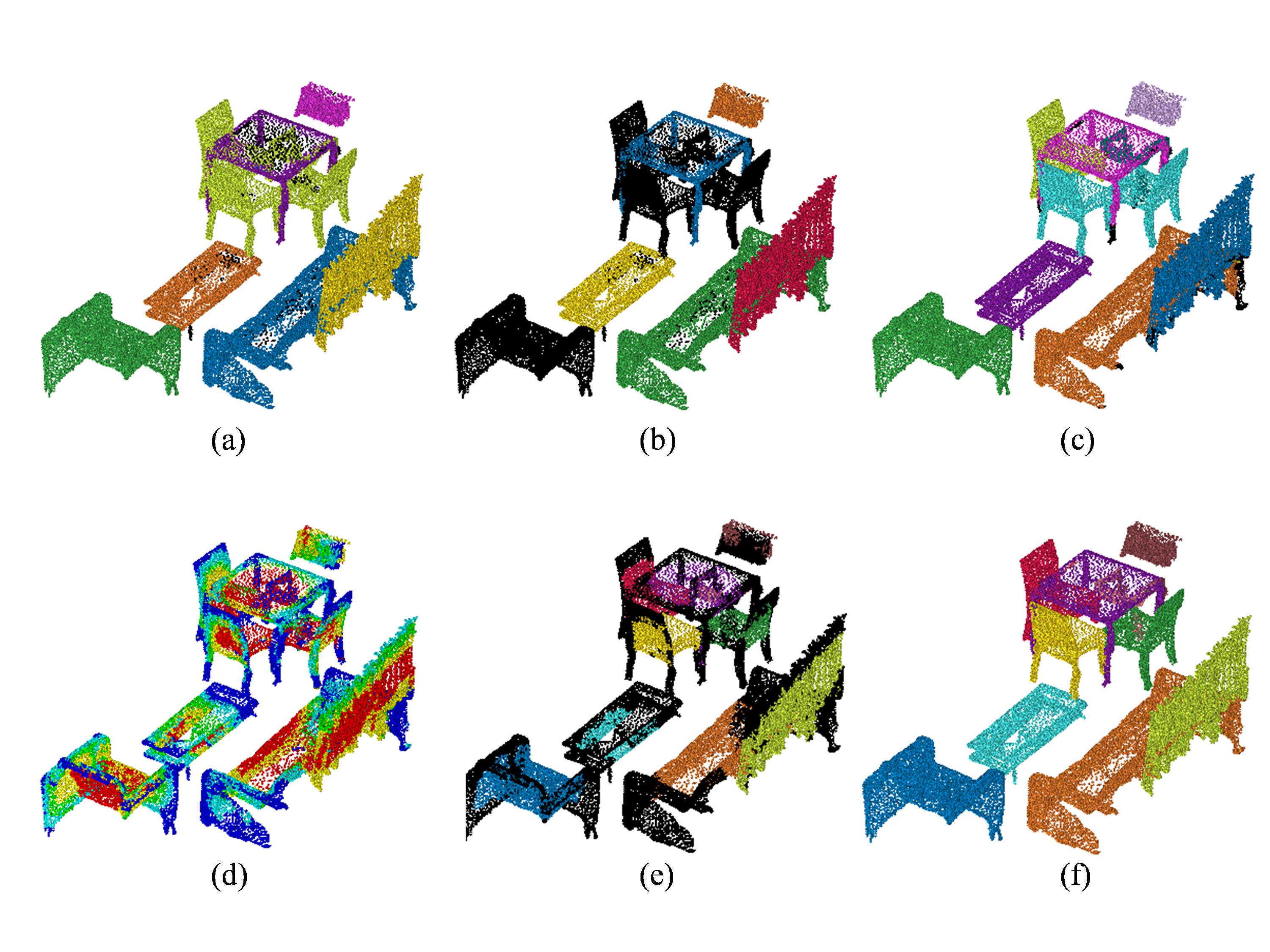}
	\end{center}
	\caption{Qualitative results of the intermediate steps in instance pseudo label generation process. (Black color points are ignored/ungrouped points.) (a) initial instance pseudo label (b) filtered instance pseudo label (c) baseline instance prediction of the network trained with filtered label (d) objectness map prediction (e) grouping results on core points (d) grouping results after absorption on boundary points}
	\label{fig:steps}
\end{figure}

\subsection{Ablation Study and Analysis}
To further analysis the effectiveness of each proposed component, we conduct following experiences.

\paragraph{Intermediate steps}
Figure \ref{fig:Pipeline} explains our workflow for generating instance pseudo labels. Here we report the visualization results on each step of the process on a training set scene in Figure \ref{fig:steps}. Given the semantic label, we use a baseline BFS algorithm on original coordinates to generate Figure \ref{fig:steps} (a). If we use these rough labels to train the network, those three chairs under the table cannot be separated. In Figure \ref{fig:steps} (b), those connected three chairs are rejected. The network trained by these selected labels can better predict offset features, which helps to separate one of the three chairs in Figure \ref{fig:steps} (c). Unlike baseline, our proposed AOIA method utilizes the objectness map in Figure \ref{fig:steps} (d).  The core points (red and yellow points) are grouped as Figure \ref{fig:steps} (e). All three chairs can then be well separated. Lastly, the boundary points (green, cyan, dark blue points) are further grouped in Figure \ref{fig:steps} (f).

\paragraph{Ablations on Optimal Sample Selection}
We conducted an experiment to verify the effectiveness of our optimal sample selection method in filtering initial pseudo labels. We trained a backbone network from scratch using two different supervisions - one with optimal sample selection and the other without. Both models were trained with the same number of epochs and settings. Our results demonstrate that the filtering strategy can improve the quality of the network's predictions by reducing the number of confusing areas during training.

\begin{table}
    \centering
    \scalebox{1.0}{
    \begin{tabular}{c|ccc}
    \toprule
    Optimal Sample Selection & \hspace{9pt}AP & \hspace{9pt}AP{\tiny 50}\hspace{9pt} & AP{\tiny 25}\\
    \midrule
    & 28.7 & 46.4 & 62.3 \\
    \checkmark & \textbf{30.3} & \textbf{48.5} & \textbf{65.5} \\
    \bottomrule
    \end{tabular}
    }
    \caption{Ablation study on the efficacy of optimal sample selection for filtering initial pseudo label. Experiments are conducted on ScanNet-v2 \cite{dai2017scannet} validation set}
    \label{tab:ablation_filtering}
\end{table}

 \begin{figure}[t]
	\begin{center}
		\includegraphics[width=1.0\linewidth]{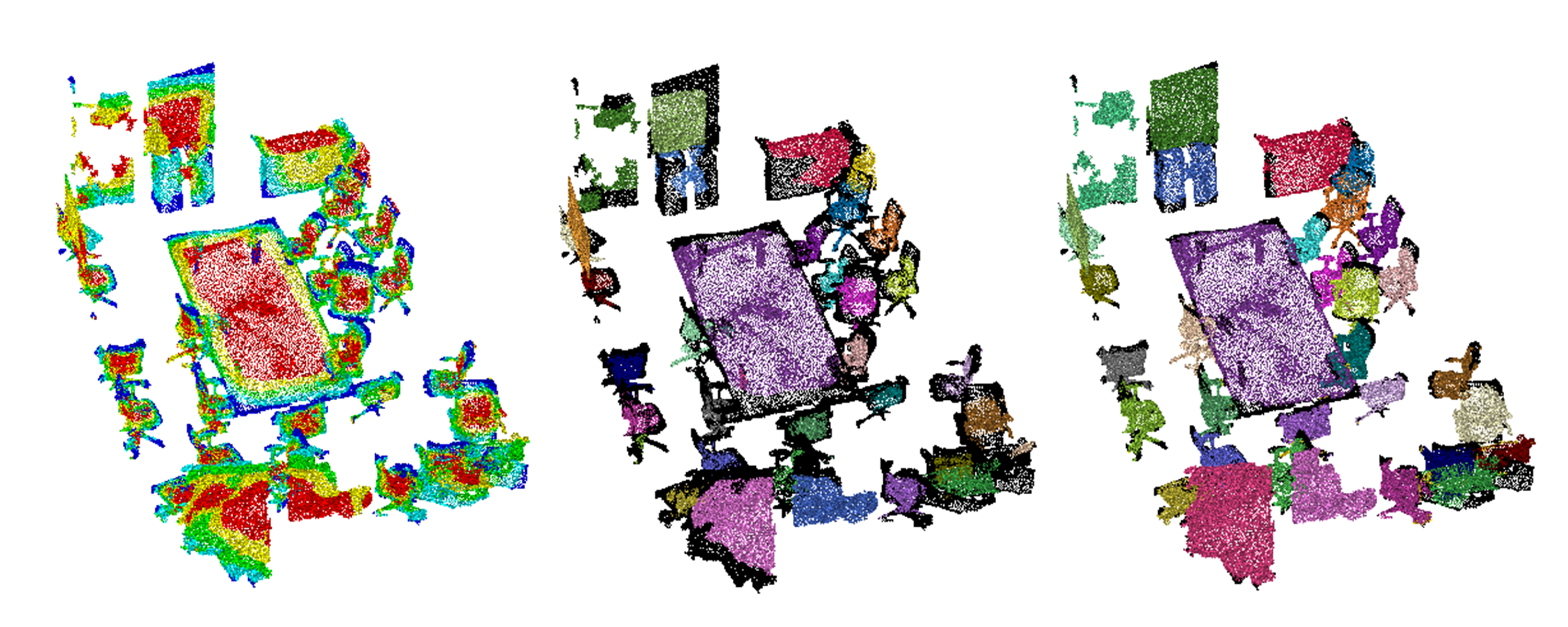}
	\end{center}
	\caption{Example of predicted objectness map and core regions by top40\% and top80\%.}
	\label{fig:objectness_map}
\end{figure}

\begin{table}
    \centering
    \scalebox{1.0}{
    \begin{tabular}{c|c|c|c|ccc}
    \toprule
    Method & Coord. & Core Pts. & Proj. & \hspace{3pt}AP & \hspace{3pt}AP{\tiny 50}\hspace{3pt} & AP{\tiny 25}\\
    \midrule
    baseline BFS & original & $100\%$ & & 62 & 73.5 & 76.9 \\
    baseline BFS & shifted & $100\%$ & & 67.9 & 78.2 & 82.1 \\
    \midrule
    AOIA (Ours) & shifted & Top$40\%$ &  & 68.5 & 78.8 & 82.9 \\
    AOIA (Ours) & shifted & Top$60\%$ &  & 69.6 & 80 & 83.7 \\
    AOIA (Ours) & shifted & Top$80\%$ &  & 70.9 & 80.8 & 83.9 \\
    AOIA (Ours) & shifted & Top$80\%$ & \checkmark & \textbf{75.6} & \textbf{83.9} & \textbf{86.3} \\
    \bottomrule
    \end{tabular}
    }
    \caption{Comparison of Instance pseudo label quality with different object inference strategies on ScanNet-v2 \cite{dai2017scannet} training set.} 
    \label{tab:ablation_algorithm}
\end{table}

\paragraph{Analysis on Asymmetric Object Inference Algorithm}
 In Table \ref{tab:ablation_algorithm} and Figure \ref{fig:objectness_map}, we analysis the inference ability of our algorithm under different settings. We use the breadth-first searching (BFS) algorithm from PointGroup \cite{pointgroup} as our baseline. Experiments are conducted on the same network after the first iteration of training. The amount of core points depends on the selection of objectness id threshold $\hat{O}$. Here we use percentages for illustration purposes. This percentage is based on the label definition, but may not be kept during prediction. The table shows using top $80\%$ as core points based on objectness maps provides a slightly better overall result. There is a trade-off in selecting the percentage of core points. In general, using less amount of core points has a stronger ability to separate instances, however it may cause over-segmentation problems in some other situations. Thus this setting is tuned for balancing the effect. In the last row, we project all point coordinates to the surface plane before them sending to the algorithm. We notice many reconstructed point cloud data are vertically broken into separate parts due to the occlusion during data collection. The bird-view effect will fairly deal with these cases and largely boost the performance. We use the projection only at the end of the first iteration. Afterward, the network can learn better shifting features with the updated pseudo label. Based on the experimental results, the projection operation is not required for the final prediction.

\paragraph{Ablations on bird view projection}
As indicated in previous paragraph, the bird view projection is proposed to deal with the hard cases during the first stage instance pseudo label generation, where point clouds are vertically broken. To further analyze the effectiveness of this operation, we compare the results of different categories in Table \ref{tab:ablation_birdview}. The results show that bird view projection is useful for most categories, except for cabinet. This meets the expectation. In real-world scenarios, most objects are not vertically stacked. Thus bird view projection can help to improve the quality of instance pseudo labels, especially when we do not have any ground-truth indication.

\begin{figure}[t]
	\begin{center}
		\includegraphics[width=1.0\linewidth]{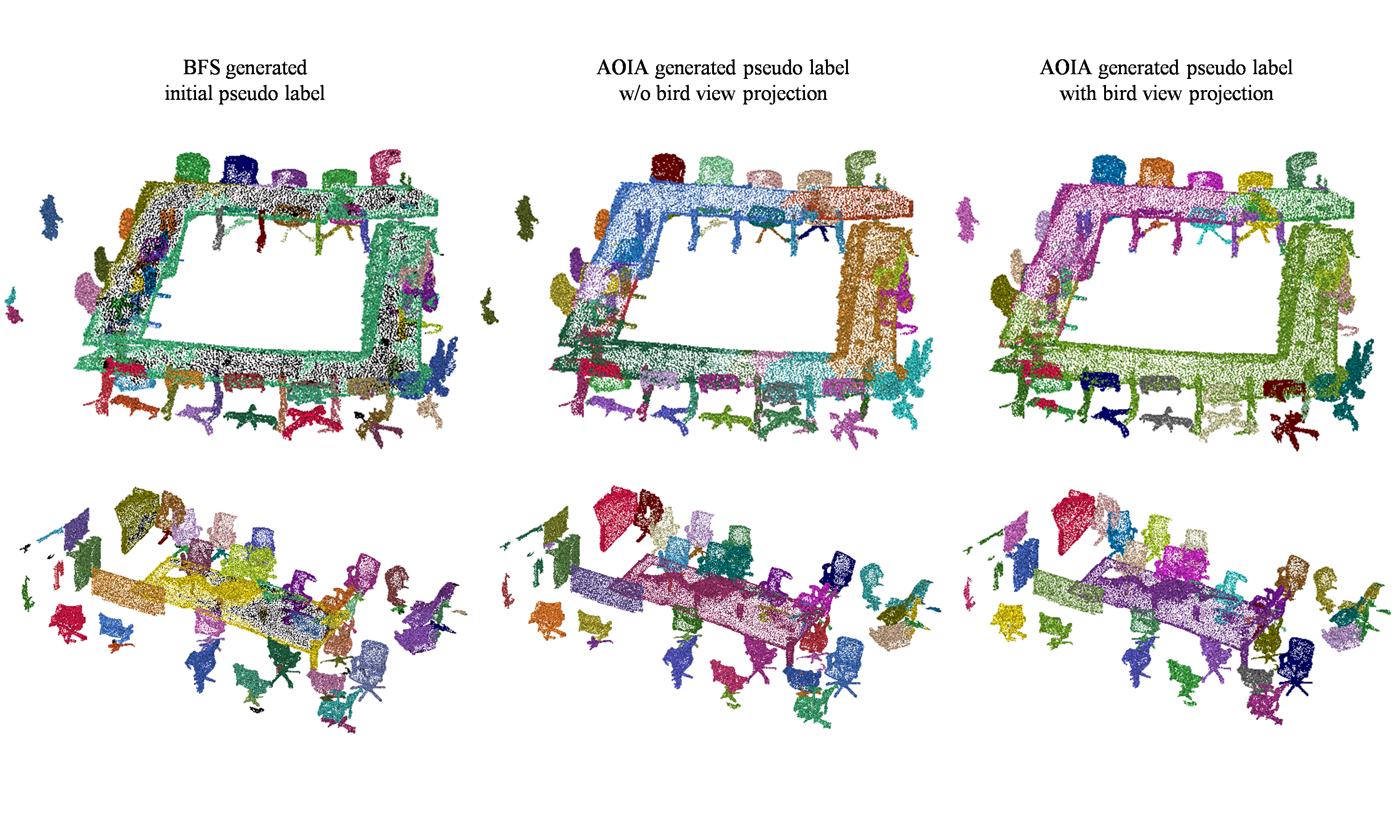}
	\end{center}
	\caption{Qualitative results of instance pseudo label with bird view projection on ScanNet v2~\cite{dai2017scannet} training set.}
	\label{fig:supp_figure4}
\end{figure}

For indoor scenes, point clouds are reconstructed from RGB-D videos. Due to the occlusion during data collection, some parts of objects can be missing. As shown in Figure \ref{fig:supp_figure4}, many point cloud objects are broken into fragments, especially for chair bases and table legs. Since they are not geometrically connected, our initial instance pseudo label generated by BFS algorithm cannot perform well. By learning shift vectors and objectness maps, our inference algorithm can better deal with these hard cases. However, some situations are still challenging. To this end, we propose to use bird view projection. The results demonstrate this operation can further help to group vertically separated point cloud instances.

\begin{figure}[t]
	\begin{center}
		\includegraphics[width=1.0\linewidth]{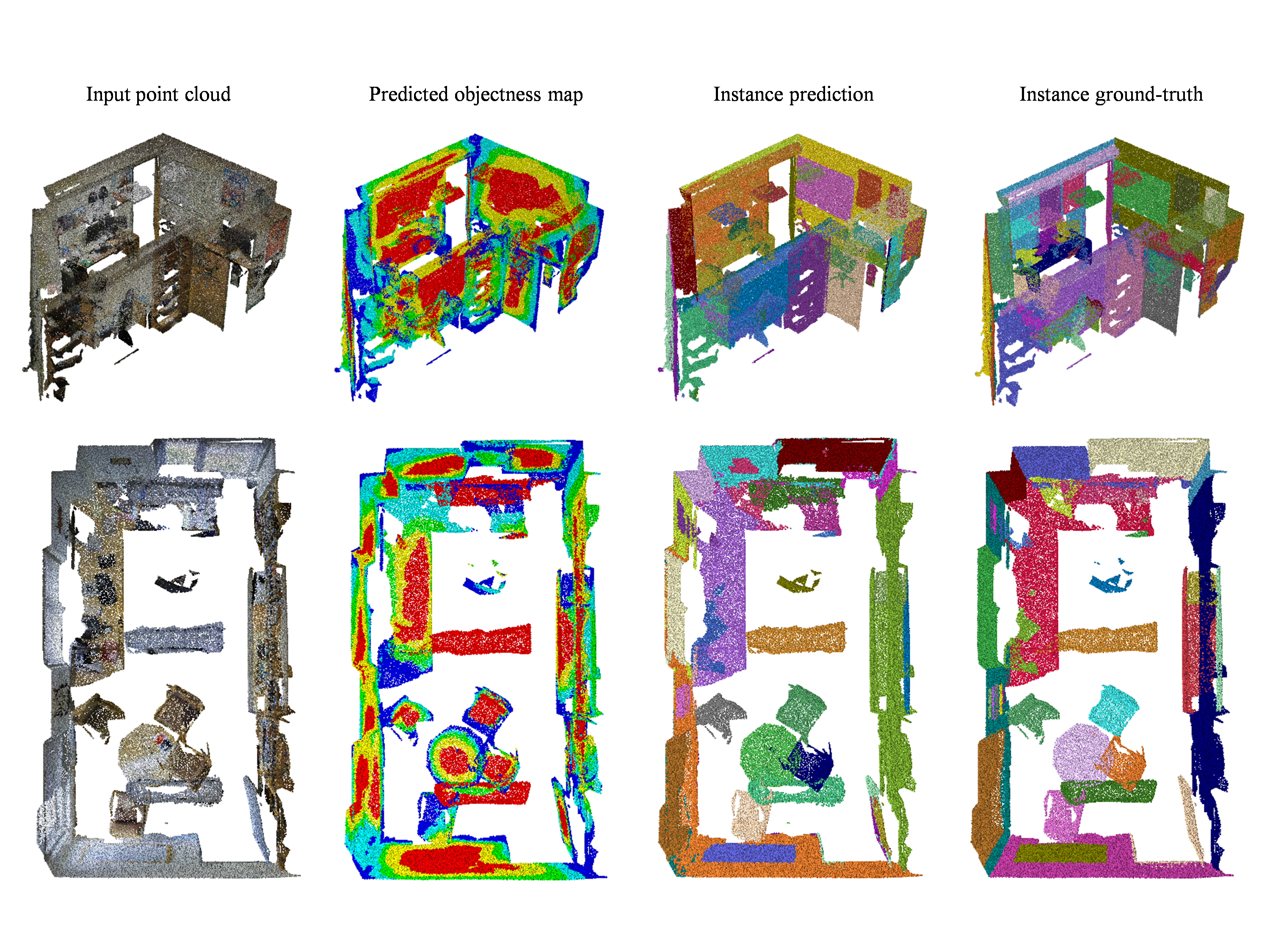}
	\end{center}
	\caption{Qualitative results of the objectness maps and instance predictions on S3DIS dataset \cite{S3DIS}}
	\label{fig:supp_figure6}
\end{figure}

\subsection{Evaluations on S3DIS Dataset}
In this section, we present the results on S3DIS dataset \cite{S3DIS}. Compared with ScanNet \cite{dai2017scannet}, S3DIS dataset \cite{S3DIS} is much smaller, containing only 272 scenes in six areas. The point cloud density is around 4 times higher than ScanNet \cite{dai2017scannet}. The configuration treats background categories (floor, ceiling, walls, etc) as instances. This is not the common way people define the concept of instance and brings some difficulty for our weakly supervised task. Following the previous methods, we evaluate the results in mean precision (mPre) and mean recall (mRec). Visualized results are shown in Figure \ref{fig:supp_figure6}. Ceilings and floors are ignored for better visualization.

\begin{table}[h]
	\begin{center}
		\scalebox{0.9}[1.0]{
			\begin{tabular}{c | c | c c | c c}
                \toprule
			     & &\multicolumn{2}{c|}{\textbf{Area 5}} & \multicolumn{2}{c}{\textbf{6-fold}}\\
				 ~~~~Method~~~~ & ~Supervision~ & ~mPrec~ & ~mRec~ & ~mPrec~ & ~mRec~ \\
                \midrule
				SGPN\cite{SGPN} & full & 36.0 & 28.7 & 38.2 & 3.12 \\
				ASIS\cite{ASIS} & full & 55.3 & 42.4 & 63.6 & 47.5 \\
				PointGroup\cite{pointgroup} & full & 61.9 & 62.1 & 69.6 & 69.2 \\
				OccuSeg\cite{occuseg} & full & - & - & 72.8 & 60.3 \\
                \midrule
				SegGroup\cite{tao2020seggroup} & weak & 47.2 & 34.9 & 56.7 & 43.3 \\
                3D-WSIS\cite{tang20223dwsis} & weak & 50.8 & 38.9 & 59.3 & 46.7 \\
				\midrule
                \textbf{AOIA (Ours)} & weak & 58.6 & 54.5 & 61.8 & 61.3 \\
				\bottomrule
			\end{tabular}
		}
	\end{center}
	\caption{3D instance segmentation results on S3DIS dataset \cite{S3DIS}}
	\label{tab:s3dis_evaluation}
\end{table}

\subsection{More Visualization Examples}
\begin{figure*}[t]
	\begin{center}
		\includegraphics[width=0.9\linewidth]{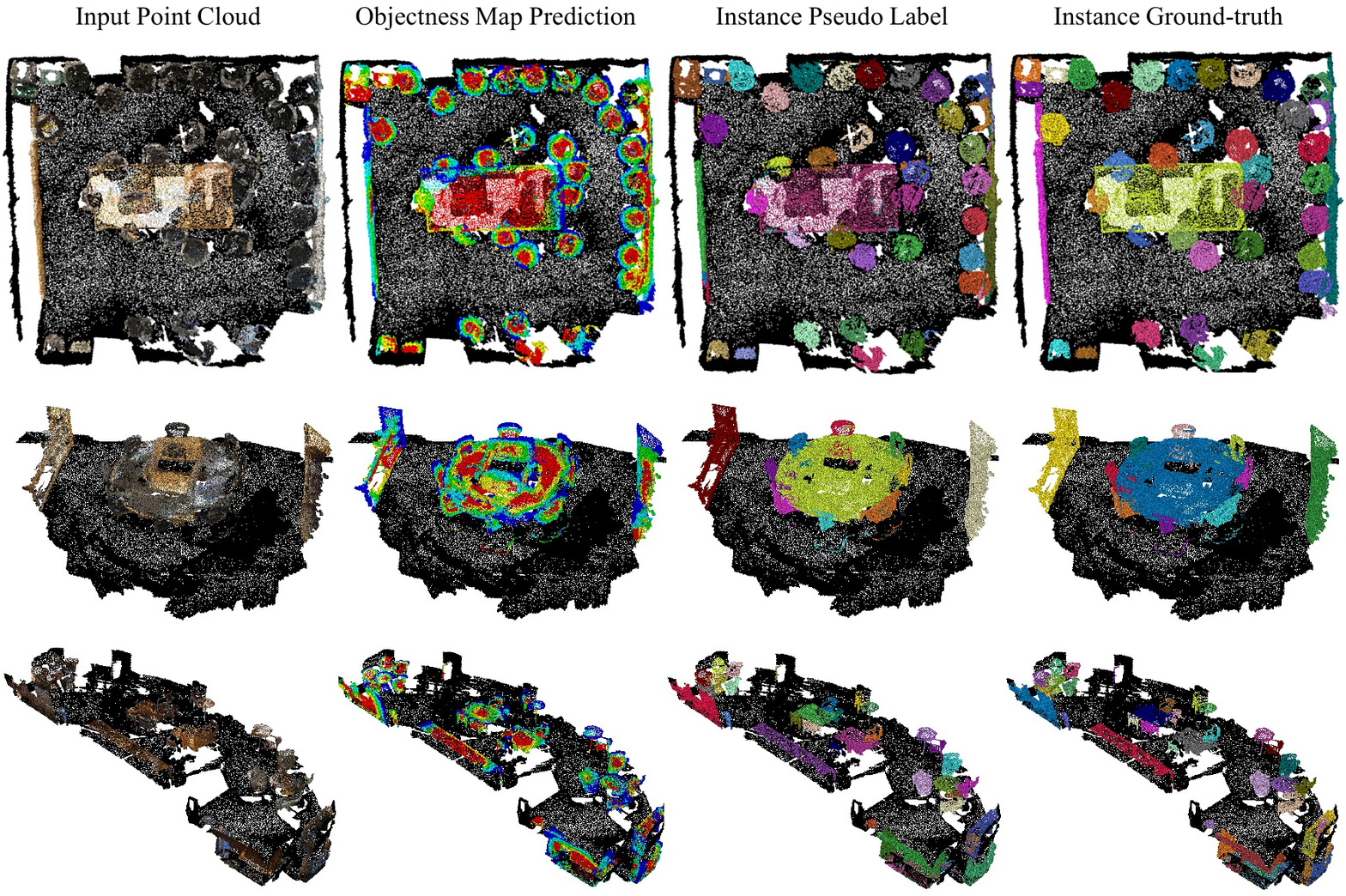}
	\end{center}
	\caption{Qualitative results of the predicted objectness maps and instance pseudo labels on ScanNet v2~\cite{dai2017scannet} training set.}
	\label{fig:supp_figure1}
\end{figure*}

We show more qualitative visualization results of our method in this section. Given raw point cloud inputs, our framework predicts the objectness maps and then generates the instance pseudo labels on ScanNet v2~\cite{dai2017scannet} training set, as shown in Figure \ref{fig:supp_figure1}. By leveraging the objectness information, the proposed asymmetric object inference algorithm may well deal with the instance pseudo label generation task.

\begin{figure*}[t]
	\begin{center}
		\includegraphics[width=0.9\linewidth]{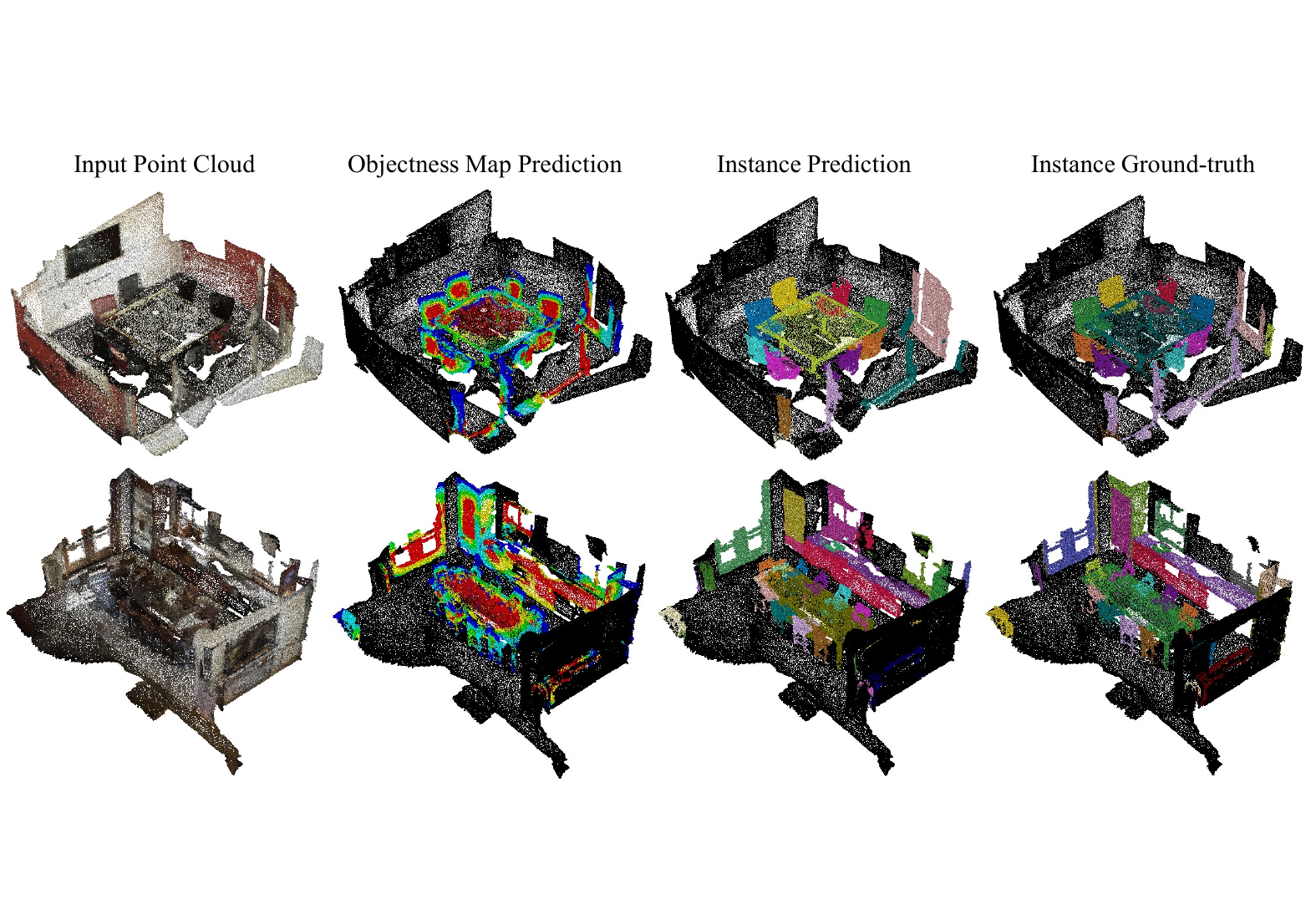}
	\end{center}
	\caption{Qualitative results of the objectness maps and instance predictions on ScanNet v2~\cite{dai2017scannet} validation set.}
	\label{fig:supp_figure2}
\end{figure*}

The predictions on validation set are shown in Figure \ref{fig:supp_figure2}. Despite our objectness network having been trained on a recomposed dataset from training set samples, it can still make fair predictions on unseen data. This shows the generalization ability of our model in predicting the objectness signals.

\begin{figure*}[t]
	\begin{center}
		\includegraphics[width=0.75\linewidth]{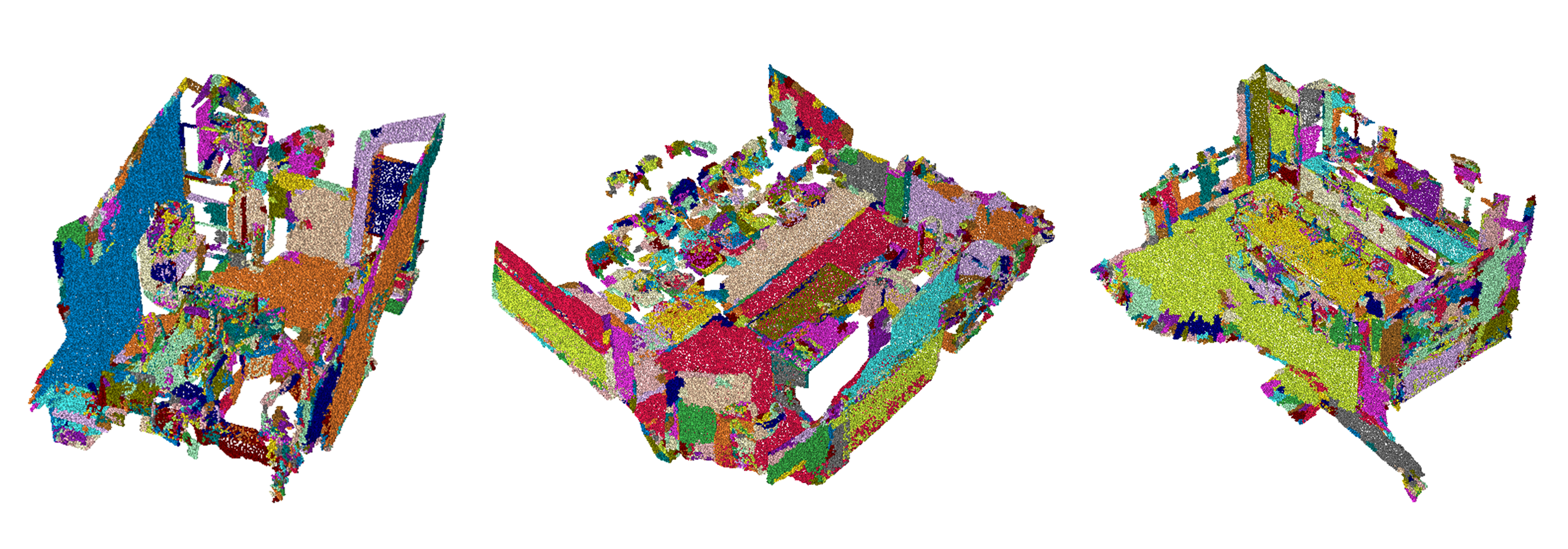}
	\end{center}
	\caption{Examples of supervoxels for over-segmentation.}
	\label{fig:supp_figure3}
\end{figure*}

In Figure \ref{fig:supp_figure3}, we show some examples of supervoxels. As introduced, we use supervoxels to smooth our semantic prediction. This is generated by the unsupervised over-segmentation method \cite{felzenszwalb2004efficient}. Since supervoxels are locally consistent, we consider points within the same supervoxel region to have the same feature by average pooling operation. Experiments show this simple label smoothing step helps to improve the performance during prediction.

\begin{figure*}[t]
	\begin{center}
		\includegraphics[width=0.75\linewidth]{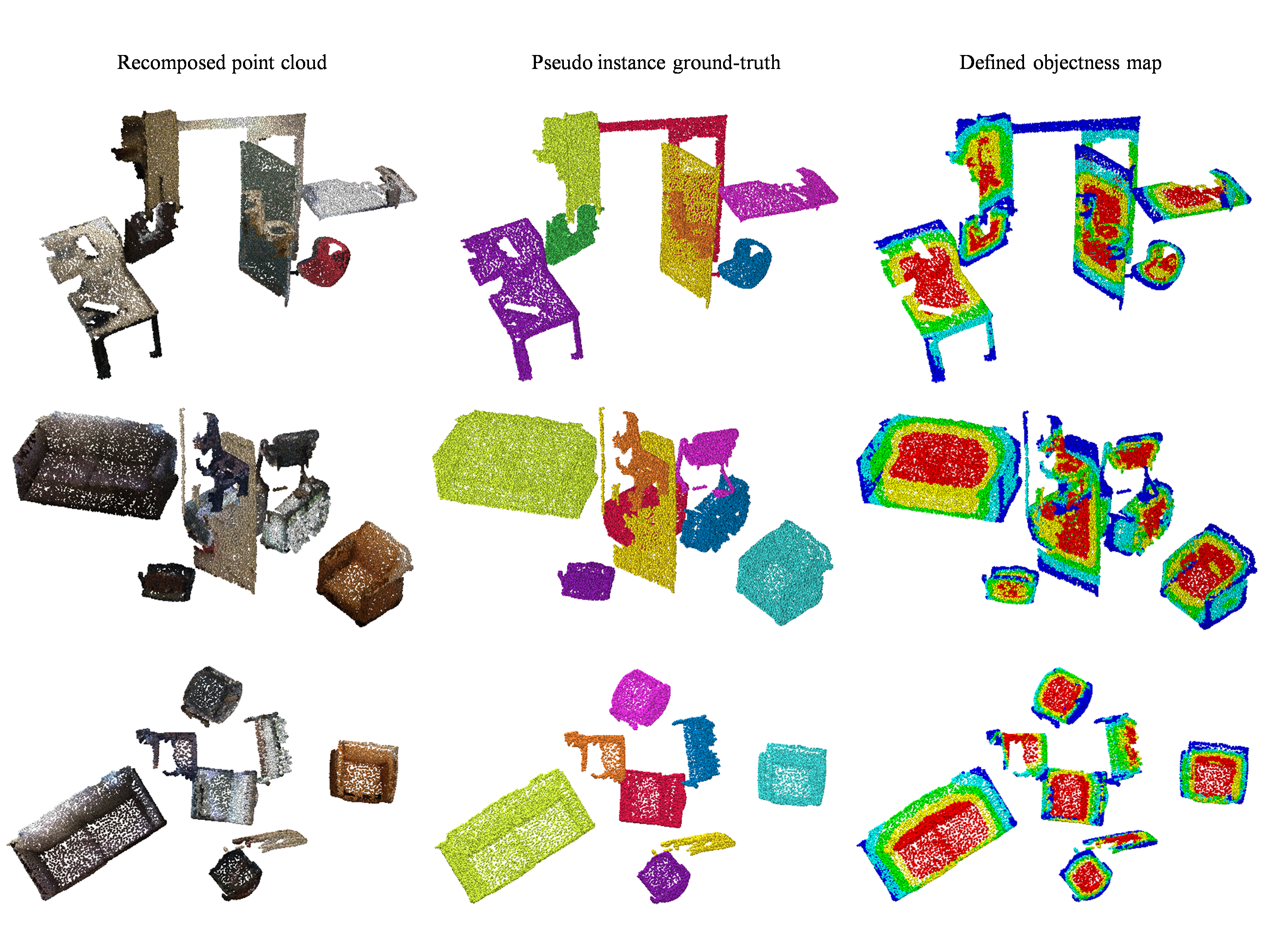}
	\end{center}
	\caption{Examples of virtual scenes in our recomposed dataset}
	\label{fig:supp_figure5}
\end{figure*}

In Figure \ref{fig:supp_figure5}, we show some sample scenes in our recomposed dataset. The scene in the third row of Figure \ref{fig:supp_figure5} is the same as the second-row scene but in the bird-view. As introduced, this dataset is randomly generated from optimal samples for our network to learn objectness signals. Specifically, we have a template that is designed on the bird-view that can contain nine samples. Each object is augmented with random rotation and may push the next object to the side if it is oversized. Our design keeps objects compactly placed and also does not have too many overlapped areas. Random operations including shifting, rotation, object drop-out are used to prevent the potential possibility of overfitting and improve the generalization ability.

\begin{table*}[t]
    \resizebox{\textwidth}{!}{
    \begin{tabular}{c|cccccccccccccccccc|c}
    \toprule
    \textbf{mAP} & cab & bed & chair & sofa & tabl & door & wind & bkshf & pic & cntr & desk & curt & fridg & showr & toil & sink & bath & ofurn & \textbf{avg}\\
    \midrule
    w/o projection&\textbf{61.7}&82.0&63.1&84.2&66.0&49.6&59.6&59.8&74.3&55.4&53.8&54.2&90.5&86.6&96.2&90.9&90.2&58.7&70.9\\
    with projection&55.3&\textbf{87.4}&\textbf{70.5}&\textbf{86.9}&\textbf{76.9}&\textbf{52.4}&\textbf{69.0}&\textbf{68.7}&\textbf{75.2}&\textbf{64.6}&\textbf{60.7}&\textbf{60.3}&\textbf{92.8}&\textbf{91.9}&\textbf{98.0}&\textbf{95.1}&\textbf{92.5}&\textbf{62.7}&\textbf{75.6}\\

    \bottomrule
    \end{tabular}
    }
    \caption{Ablation on bird view projection for instance pseudo label generation on ScanNet v2~\cite{dai2017scannet} training set.}
    \label{tab:ablation_birdview}
\end{table*}

\subsection{Analysis on Objectness Signal}
\begin{figure*}[t]
	\begin{center}
		\includegraphics[width=0.75\linewidth]{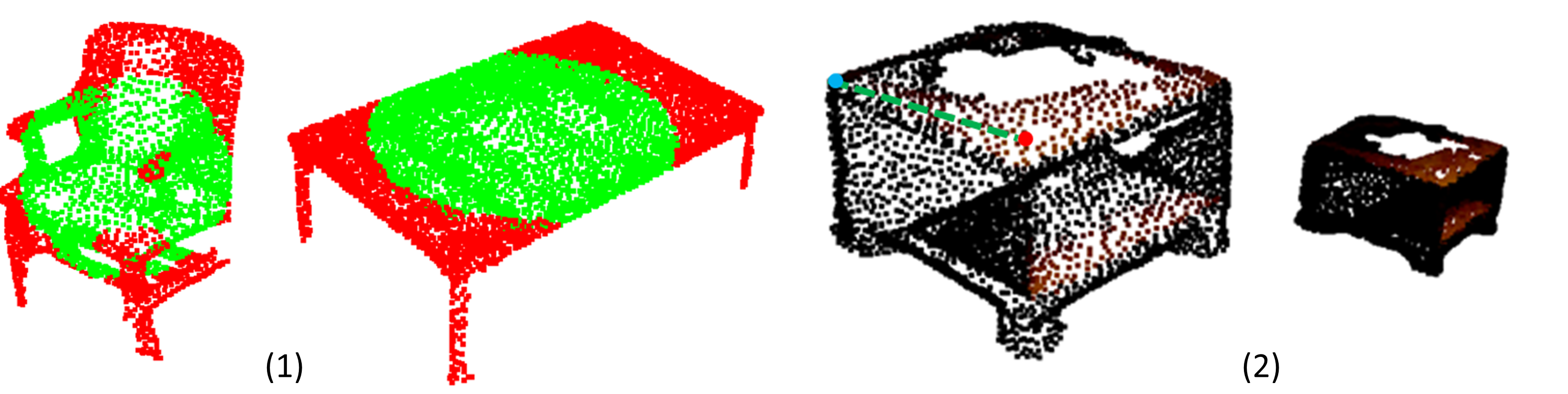}
	\end{center}
	\caption{(1) Objectness based on the pure distance from the centroid (2) Sort the points by the nearest distances to the compressed object rather than a single centroid (our method)}
	\label{fig:objectness_example}
\end{figure*}

We have shown in Section \ref{sec:objectness_branch} and Figure \ref{fig:Toydata} on how to generate the defined objectness signals. In this section, we provide further details on our approach for defining objectness signals and present negative examples to illustrate the limitations of alternative methods. We begin by considering a simple and naive approach that relies on calculating the pure distance to the object center and selecting the top $50\%$ of the nearest points as core points. However, the use of a pure distance based approach to measure objectness lacks shape adaptivity and may lead to inaccuracies when applied to objectness-guided clustering. As depicted in Figure \ref{fig:objectness_example} (1), the green points always form a spherical pattern, and using this objectness signal may produce erroneous clusters if the object's shape deviates from a sphere.

To overcome this issue, we propose an alternative approach that utilizes a compressed point cloud as a reference (shown in Figure \ref{fig:objectness_example} (2)). Instead of measuring the distance between a point and its corresponding object centroid point, we calculate the nearest distance from a point to the compressed point cloud. Please note that, in our rule design, the compressed point cloud is located inside the original point cloud, and both point clouds share the same centroid point.

\section{Conclusion}
In this paper, we have proposed a novel weakly supervised 3D instance segmentation method that requires no instance-level annotations. We explore the in-between space between point clouds and find those clusters with a high likelihood of being complete objects. Then, we build a randomly recomposed dataset with those selected samples for learning our defined objectness signal, which encodes instance-aware information. We further design an asymmetric object inference algorithm for effective instance pseudo label generation. Experimental results showed our approach can produce high-quality instance pseudo labels and achieve comparable performance with the fully supervised baseline method.

\section*{Acknowledgments}
This study is supported under the RIE2020 Industry Alignment Fund – Industry Collaboration Projects (IAF-ICP) Funding Initiative, as well as
cash and in-kind contribution from the industry partner(s).
This research is partly supported by the MoE AcRF Tier 2 grant (MOE-T2EP20220-0007) and the MoE AcRF Tier 1 grant (RG14/22).

\ifCLASSOPTIONcaptionsoff
  \newpage
\fi

\bibliographystyle{IEEEtran}
\bibliography{egbib}
\end{document}